\documentclass[hidelinks]{isprs}
\usepackage{subfigure}
\usepackage{import}
\usepackage{setspace}
\usepackage{geometry} % added 27-02-2014 Markus Englich
\usepackage{epstopdf}
\usepackage[labelsep=period]{caption}  % added 14-04-2016 Markus Englich - Recommendation by Sebastian Brocks

\usepackage{hyperref}
\usepackage{graphicx}
\usepackage[export]{adjustbox}
\usepackage{amsmath,amsfonts,amssymb}
\usepackage[acronyms]{glossaries}
\usepackage[sort]{natbib} % for \citet
	\renewcommand{\refname}{REFERENCES} % natbib turns Refernces to lower case.
\usepackage{color, soul}
\usepackage{diagbox}	% diagonal box
\usepackage{enumitem}
\usepackage{gensymb}
\usepackage[nameinlink, capitalize]{cleveref}
\usepackage{hhline}
\usepackage{inputenc}

\usepackage[percent]{overpic}

\DeclareMathOperator*{\argmin}{arg\,min}

\usepackage{xspace}

\newcommand{\ie}{i.e.\ }

\newcommand{\eg}{e.g.\ }

\newcommand{\cf}{cf.\ }

\newcommand{\m}{\,m\xspace}

\newcommand{\fps}{\,fps\xspace}

\newcommand{\xSGM}{SGM$^x$\xspace}

\newcommand{\fpSGM}{SGM$^{fp}$\xspace}

\newcommand{\snSGM}{SGM$^{sn}$\xspace}

\newcommand{\pgSGM}{SGM$^{pg}$\xspace}

\begin{document}

\title{Efficient Surface-Aware Semi-Global Matching with Multi-View Plane-Sweep Sampling}
%\title{Efficient Multi-View Semi-Global Stereo Matching in Object Space with Surface Normal Smoothing and Edge Enhancement}

% Leave Blank for double blind submission
\author{
B. Ruf\textsuperscript{a,b}, T. Pollok\textsuperscript{a}, M. Weinmann\textsuperscript{b}
}

% Leave blank for double blind submission
\address{
\textsuperscript{a}Fraunhofer IOSB, Video Exploitation Systems, Karlsruhe, Germany -\\ \{boitumelo.ruf, thomas.pollok\}@iosb.fraunhofer.de\\
\textsuperscript{b}Institute of Photogrammetry and Remote Sensing, Karlsruhe Institute of Technology, \\Karlsruhe, Germany - \{boitumelo.ruf, martin.weinmann\}@kit.edu
}

% If the corresponding author is NOT the final author, always add a % space before the subsequent comma, i.e.
% first author name\textsuperscript{a,}\thanks{Corresponding author} , % second author name \textsuperscript{b}, etc.
% thanks to Niclas Borlin 05-05-2016

\commission{II, }{II} %This field is optional.
\workinggroup{II/4} %This field is optional.
\icwg{}   %This field is optional.


\newacronym{ASG}{ASG}{{Average Shading Gradient}}

\newacronym[shortplural={CNNs}, firstplural={convolutional neural networks}, longplural={convolutional neural networks}]{CNN}{CNN}{convolutional neural network}
\newacronym{COTS}{COTS}{commercial off-the-shelf}
\newacronym{CT}{CT}{{Census Transform}}

\newacronym{DLT}{DLT}{{Direct Linear Transformation}}
\newacronym{DoG}{DoG}{{Difference of Gaussian}}

\newacronym{EPnP}{EPnP}{{Efficient Perspective-n-Point}}

\newacronym{GPGPU}{GPGPU}{general purpose computation on a {GPU}}
\newacronym{GPS}{GPS}{{Global Positioning System}}

\newacronym{ICP}{ICP}{{Iterative-Closest-Point}}
\newacronym{IMU}{IMU}{{Inertial Measurement Unit}}
\newacronym{INS}{INS}{{Inertial Navigation System}}

\newacronym{LIDAR}{LiDAR}{light detection and ranging}

\newacronym[shortplural={MRFs}, longplural={{Markov Random Fields}}]{MRF}{MRF}{{Markov Random Field}}
\newacronym{MVS}{MVS}{Multi-View Stereo}

\newacronym{NCC}{NCC}{normalized cross correlation}

\newacronym{PCL}{PCL}{{Point Cloud Library}}

\newacronym{RANSAC}{RANSAC}{{Random Sampling Consensus}}
\newacronym[shortplural={ROIs}, longplural={regions of interest}]{ROI}{RoI}{region of interest}

\newacronym{SAD}{SAD}{sum of absolute differences}
\newacronym{SFM}{SfM}{{Structure-from-Motion}}
\newacronym{SGM}{SGM}{{Semi-Global Matching}}

\newacronym[shortplural={UAVs}, longplural={unmanned aerial vehicles}]{UAV}{UAV}{unmanned aerial vehicle}

\newacronym{WTA}{WTA}{winner-takes-it-all}

% KAO: Use times symbol
\abstract{
Online augmentation of an oblique aerial image sequence with structural information is an essential aspect in the process of 3D scene interpretation and analysis. %
One key aspect in this is the efficient dense image matching and depth estimation. %
Here, the \gls*{SGM} approach has proven to be one of the most widely used algorithms for efficient depth estimation, providing a good trade-off between accuracy and computational complexity. %
However, \gls*{SGM} only models a first-order smoothness assumption, thus favoring fronto-parallel surfaces. %
In this work, we present a hierarchical algorithm that allows for efficient depth and normal map estimation together with confidence measures for each estimate. %
Our algorithm relies on a plane-sweep multi-image matching followed by an extended \gls*{SGM} optimization that allows to incorporate local surface orientations, thus achieving more consistent and accurate estimates in areas made up of slanted surfaces, inherent to oblique aerial imagery. %
We evaluate numerous configurations of our algorithm on two different datasets using an absolute and relative accuracy measure. %
In our evaluation, we show that the results of our approach are comparable to the ones achieved by refined \gls*{SFM} pipelines, such as COLMAP, which are designed for offline processing. %
In contrast, however, our approach only considers a confined image bundle of an input sequence, thus allowing to perform an online and incremental computation at 1Hz$-$2Hz.
%Our approach allows for an online structural scene analysis from oblique aerial imagery, if joined with the online estimation of camera poses and simultaneous localization with respect to a reference system.
}

\maketitle

\glsresetall % reset all acronyms
%\glsunset{UAV} % unset UAV acronym as UAVs is used firsts

%%%%%%%%%%%%%%%%%%%%%%%%%%%%%%%%%%%%%%%%%%%%%%%
\section{INTRODUCTION}
\label{sec:intro}
%%%%%%%%%%%%%%%%%%%%%%%%%%%%%%%%%%%%%%%%%%%%%%%
% KAO: Sloppy spacing ensures non-overfull lines. Can be removed if this is not an issue.
\sloppy

Dense image matching is one of the most important and intensively studied task in photogrammetric computer vision. %
It allows to estimate dense depth maps which, in turn, alleviate the processes of dense 3D reconstruction and model generation \citep{Musialski2013, Rothermel2014, Blaha2016, Bulatov2011multi}, navigation of autonomous vehicles such as robots, cars and \glspl*{UAV} \citep{Menze2015CVPR, Scaramuzza2014, Barry2015}, as well as scene interpretation and analysis \citep{Taneja2015Change, Weinmann2016}. %
Especially in combination with small \gls*{COTS} \glspl*{UAV} it allows for a cost-effective monitoring of man-made structures from aerial viewpoints. %

In general, dense image matching algorithms can be grouped into two categories, namely \textit{local} and \textit{global} methods \citep{Scharstein2002}. %
Since local methods only consider a confined neighborhood by aggregating a matching cost in a local aggregation window, they can be computed very efficiently, allowing to achieve real-time processing. %
However, their smoothness assumptions are restricted to the local support region and therefore the accuracies achieved by these methods are typically not in the order of those achieved by global methods. %

First introduced by \citet{Hirschmueller2005, Hirschmueller2008}, \gls*{SGM} combines the benefits of both local and global methods. %
The use of dynamic programming to approximate the energy minimization, by independently aggregating along numerous concentric one-dimensional paths, provides a good trade-off between accuracy and computational complexity. %
Thus, \gls*{SGM} is still one of the most widely used algorithms for efficient image-based depth estimation from both two-view and multiple-view setups. %
Furthermore, recent studies show that the \gls*{SGM} algorithm can be adapted to allow for real-time stereo depth estimation solely on a desktop CPU \citep{Gehrig2010, Spangenberg2014} or embedded hardware \citep{Banz2011, Hofmann2016, Ruf2018embedded}. %

However, \gls*{SGM} only models a first-order smoothness assumption, thus favoring fronto-parallel surfaces. %
This is sufficient for applications, in which the existence of a reconstructed 3D object is more important than its detailed appearance, such as robot navigation. %
Nonetheless, when it comes to a visually accurate 3D reconstruction of slanted surfaces, which are inherent to oblique aerial imagery, a second-order smoothness assumption is desirable. %
To overcome this restriction, \citet{Scharstein2018surface} propose to incorporate priors, such as normal maps, to dynamically adjust \gls*{SGM} to the surface orientation of the object that is to be reconstructed. %

In this work, we propose an algorithm that extends \gls*{SGM} to a multi-image matching, which allows for online augmentation of an aerial image sequence with structural information and focuses on oblique imagery captured from small \glspl*{UAV}.
Thus, our contribution is an approach for image-based depth estimation, that  %
\begin{itemize} %
\item relies on a hierarchical multi-image semi-global stereo matching, %
\item favors not only fronto-parallel surfaces but incorporates a regularization based on local surface normals, and
%\item enhances depth discontinuities at object boundaries, and %
\item allows for efficient depth and normal map estimation with confidence measures from aerial imagery. %
\end{itemize} %
This paper is structured as follows: In \Cref{sec:related_work}, we briefly summarize the related work on algorithms that rely on \gls*{SGM} and allow for efficient image-based depth estimation. %
We specifically focus on the use of non-fronto-parallel smoothness assumptions allowing for slanted surface reconstruction. %
In \Cref{sec:methodology}, we give a detailed overview on our methodology, focusing on our adaptation of \gls*{SGM} to be used with multi-image matching for dense depth estimation from oblique aerial imagery, together with the estimation of surface normals and confidence measures. %
We evaluate our approach on two datasets (\Cref{sec:eval}) and present our achieved results in \Cref{sec:experiments}, which is followed by a discussion in \Cref{sec:discussion}. %
Finally, we provide a summary, concluding remarks, and a short outlook on future improvements in \Cref{sec:conclusion}. %

%%%%%%%%%%%%%%%%%%%%%%%%%%%%%%%%%%%%%%%%%%%%%%%
\section{RELATED WORK}%
\label{sec:related_work}
%%%%%%%%%%%%%%%%%%%%%%%%%%%%%%%%%%%%%%%%%%%%%%%
% KAO: Sloppy spacing ensures non-overfull lines. Can be removed if this is not an issue.
\sloppy
\glsreset{SGM}

In recent years, a number of software suites to address accurate dense 3D reconstruction have been released. %
These include the \gls*{SFM} pipelines SURE \citep{Rothermel2012,Wenzel2013Sure} and COLMAP \citep{Schoenberger2016mvs, Schoenberger2016sfm}, that enable the creation of detailed 3D models from a large set of input images. %
While the focus of these pipelines lies on the accuracy and completeness of the resulting 3D model, they are designed for offline processing, in which computation time is not a critical factor and all input images are available at the time of reconstruction. %
However, since our work focuses on online processing, computation time is a critical factor for us and we cannot assume that the complete input sequence is available for the process of 3D reconstruction. %
In addition, since we aim at methods that generate a dense field of depth estimates, we use a direct dense image matching for the computation of depth maps, instead of sparse feature matching. %

When it comes to efficient dense image matching, the \gls*{SGM} algorithm \citep{Hirschmueller2005, Hirschmueller2008} has evolved to a suitable and widely used approach. %
The accuracy achieved with respect to the computation time needed makes \gls*{SGM} very appealing for both offline and online processing. %
In their work, \citet{Spangenberg2014} as well as \citet{Gehrig2010} show that, when using a fixed stereo setup, \gls*{SGM} can be optimized to run at 16\fps and 14\fps, respectively, on a conventional desktop CPU when utilizing SIMD instructions and using input images at VGA resolution. %
The most common optimization strategy for the \gls*{SGM} algorithm, however, is to utilize the massively parallel computation infrastructure of modern GPUs, achieving real-time frame rates \citep{Banz2011}. %
An alternative is to allow for dense image matching from aerial imagery with large disparities by encapsulating the \gls*{SGM} approach in a hierarchical processing scheme \citep{Rothermel2012, Wenzel2013Sure}. %
Even in the field of embedded stereo processing, real-time performance with high accuracies can be achieved by optimizing the \gls*{SGM} approach for FPGA architectures \citep{Hofmann2016, Barry2015, Ruf2018embedded}. %
%As a thorough analysis of all approaches that rely on \gls*{SGM} is beyond the scope of this discussion, we provide a short excerpt of noteworthy applications. % 

In more recent work, \citet{Scharstein2018surface} have proposed an improvement to the accuracy of the \gls*{SGM} approach by including available surface priors to better cope with slanted surfaces and untextured regions. %
Similarly, \citet{Hermann2009inclusion} and \citet{Ni2018second} propose to extend \gls*{SGM} by incorporating a second-order smoothness assumption, that also allows to favor non-fronto-parallel surfaces. %

The so-called plane-sweep sampling for true multi-image matching was first introduced by \citet{Collins1996space} and was adopted in a great amount of studies aiming at real-time depth estimation and 3D reconstruction from image sequences. 
Among many are the work of \citet{Gallup2007} and \citet{Sinha2014}. 
\citet{Gallup2007} introduced an extension to the plane-sweep algorithm that does not only consider a fronto-parallel sweeping direction but also incorporates other plane orientations that align with the scene geometry, \eg ground plane.
\citet{Sinha2014} further extend the plane-sweep approach for multi-image matching by identifying different plane configurations for local image regions in contrast to using the same plane orientations for the whole image.
Furthermore, \citet{Sinha2014} also propose to use the semi-global optimization strategy to extract the final disparity image from the result of the local plane-sweep sampling.

In our work, we incorporate and evaluate the strengths of multiple approaches by using a hierarchical multi-view image matching and considering surface normals to better handle non-fronto-parallel surfaces in the semi-global optimization scheme. %

%%%%%%%%%%%%%%%%%%%%%%%%%%%%%%%%%%%%%%%%%%%%%%%
\section{METHODOLOGY}%
\label{sec:methodology}
%%%%%%%%%%%%%%%%%%%%%%%%%%%%%%%%%%%%%%%%%%%%%%%
% KAO: Sloppy spacing ensures non-overfull lines. Can be removed if this is not an issue.
\sloppy

\begin{figure}[htb]%
	\centering%
	\resizebox{\columnwidth}{!}{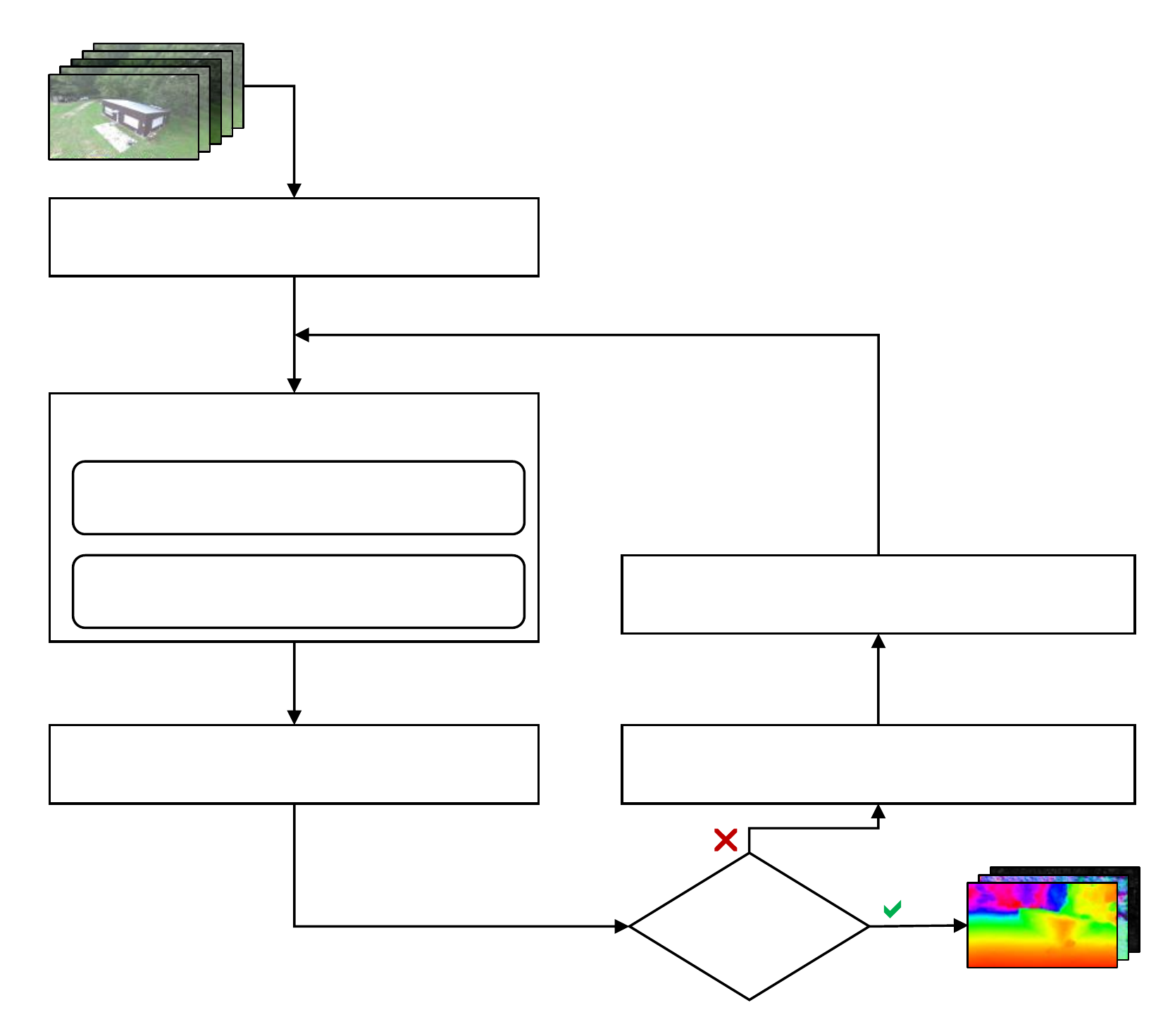}%
	\caption{%
		\footnotesize{
		Overview of the proposed methodology. %
		Given five images $I_i$ of an input sequence, we perform hierarchical \gls*{SFM} to estimate a depth, confidence and normal map ($\hat{D}, \hat{C}, \hat{N})$.
		}
	}%
	\label{fig:methodology}%
\end{figure}

\Cref{fig:methodology} depicts the processing pipeline of our approach for a hierarchical multi-image matching followed by a surface-aware and edge-preserving \gls*{SGM} optimization together with a computation of surface normals and confidence measures. %
We first give a brief overview on all processing steps before we provide a detailed description of our extensions of the \gls*{SGM} algorithm, the computation of confidence measures, as well as a detailed explanation on the computation of surface normals. %

As input to our processing pipeline, we choose a bundle of five images $I_i$ of an input sequence which depict the scene that is to be reconstructed from five slightly different viewpoints. %
We select the center image of the input bundle as reference image $I_\mathrm{ref}$ for which the depth, normal and confidence maps are to be computed.
To this end, we assume that the camera poses $\mathbf{P}_i = \mathbf{K} \left[\mathbf{R}_i^\mathrm{T}\ \ -\mathbf{R}_i^\mathrm{T} \mathrm{C}_i \right]$ are given together with the input images. %
Here, $\mathrm{C}_i \in \mathbb{R}^3$ and $\mathbf{R}^\mathrm{T}_i \in SO(3)$ denote the locations of the camera centers and the orientations of the cameras relative to a given reference coordinate system. %
Since we assume the input images to belong to the same image sequence, we set the intrinsic calibration matrix $\mathbf{K}$ equal for all images.

Given the input bundle, we first compute image pyramids for each input image, which allow for a hierarchical processing in the subsequent steps. %
Assuming that the lowest level of each pyramid is the original input image, we perform a Gaussian blurring, with $\sigma = 1$ and a $3\times 3$ kernel, before reducing the image size of each pyramid level by a factor of two in both image directions with respect to the previous level when successively moving up the pyramid. %
This yields a bundle of five image pyramids corresponding to the five input images with $L$ pyramid levels. %
We initialize our algorithm to start off at the coarsest pyramid level $l = 0, l \in L$, \ie the level with the smallest image dimensions, and a full sampling range between $\left[d^{\,0}_{\,\mathrm{min}}, d^{\,0}_{\,\mathrm{max}}\right]$. %

For each pyramid level $l$, we first compute a three-dimensional matching cost volume $\mathcal{S}$ by employing a real-time plane-sweep multi-image matching \citep{Collins1996space}, sampling the scene space for each pixel $\mathrm{p}$ between two fronto-parallel bounding planes $\Pi_{\,\mathrm{p,\,max}}$ and $\Pi_\mathrm{\,\mathrm{p,\,min}}$ located at $d^{\,l}_{\,\mathrm{p,\,max}}$ and $d^{\,l}_{\,\mathrm{p,\,min}}$ respectively. %
For this, we adopt the approach presented by \citet{Ruf2017cross} to select the set of sampling planes such that, when considering one of the corner pixels in $I_\mathrm{ref}$, two consecutive planes cause a maximum pixel displacement of 1 on the corresponding epipolar line in the matching image with the largest distance to $I_\mathrm{ref}$. %
Note that the parameters for plane-induced homographies are different for each pyramid level, since the change in image size affects the intrinsic camera matrix $\mathbf{K}$. %
Furthermore, at the first pyramid level, we sample within the full sampling range for each pixel, while adopting a pixel-wise sampling range in the successive steps according to the previously predicted depth map. %
 
As a similarity measure for the multi-image matching, we use and evaluate the Hamming distance of the \gls*{CT} \citep{Zabih1994}, as well as a negated, truncated and scaled form of the \gls*{NCC} as described in \citep{Scharstein2018surface}. %
To account for occlusions, we adopt the approach presented by \citet{Kang2001}, selecting the minimum aggregated matching costs of the left and right subset with respect to the reference image $I_\mathrm{ref}.$ 
The resulting cost volume $\mathcal{S}^l$ corresponding to pyramid level $l$ is of size $\mathcal{W}^l\times \mathcal{H}^l\times \mathcal{D}^l$, where $\mathcal{W}^l$ and $\mathcal{H}^l$ denote the image size and $\mathcal{D}^l$ is the number of planes with which the scene is sampled at the current pyramid level. %
Since the per-pixel sampling range at pyramid levels $l>0$ differ, we employ a dynamic cost volume \citep{Wenzel2013Sure}.

In the next step, $\mathcal{S}^l$ is regularized with a semi-global optimization scheme, yielding a dense depth map $D^l$ together with pixel-wise confidence measures of the estimated depth stored in a confidence map $C^l$. %
In this work, we propose three different optimization schemes (\xSGM) that extend the \gls*{SGM} approach initially presented by \citet{Hirschmueller2005, Hirschmueller2008}. %
These include a straight-forward extension used together with a plane-sweep sampling favoring fronto-parallel surfaces \citep{Ruf2017cross}, as well as adopting the approach of \citet{Scharstein2018surface} to use available surface normal information in order to also favor slanted surfaces. %
Furthermore, we incorporate two strategies to adapt the penalties of the smoothness term of \gls*{SGM}, thus preserving edges at object boundaries in the depth maps. % 
A detailed description on our \gls*{SGM} optimization and the confidence measures used can be found in \Cref{sec:xsgm} and \Cref{sec:conf}. %

Given the estimated depth map, we compute a normal map $N^l$, which is not only an additional output of our algorithm but is also needed to adapt our \xSGM optimization to the surface normals in the following iteration of our hierarchical processing scheme. %
Here, we employ an appearance-based weighted Gaussian smoothing, which we call \textit{Gestalt-Smoothing}, that regularizes the normal map while preserving discontinuities based on the appearance between neighboring pixels. %
Details on our approach for the extraction of surface normals from a single depth map are given in \Cref{sec:normalmap}.

If the lowest level of the image pyramids has not yet been reached within our hierarchical processing envelope, we use the depth map $D^l$ and normal map $N^l$ to initialize and regularize the depth map estimation at the next pyramid level $l+1$. %
In doing so, $D^l$ and $N^l$ are first upscaled with nearest neighbor interpolation to the image size of the next pyramid level, yielding $\bar{D}^l$ and $\bar{N}^l$. %
The upscaled depth map is used to reinitialize the homography-based plane-sweep sampling by restricting the sampling range for each pixel $\mathrm{p}$. %
For this, we use the predicted depth value $\bar{d}^{\,l}_{\,\mathrm{p}} = \bar{D}^{\,l}(\mathrm{p})$ and set the per-pixel sampling range to $\left[d^{\,l+1}_{\,\mathrm{p,\,min}} = \bar{d}^{\,l}_{\,\mathrm{p}}-\Delta d, d^{\,l+1}_{\,\mathrm{p,\,max}} = \bar{d}^{\,l}_{\,\mathrm{p}}+\Delta d\right]$. %
Since the homographic mappings are precomputed, we select the per-pixel bounding planes $\Pi_{\,\mathrm{p, max}}^{\,l+1}$ and $\Pi_{\,\mathrm{p, min}}^{\,l+1}$ as the closest planes to $d^{\,l+1}_{\,\mathrm{p,\,max}}$ and $d^{\,l+1}_{\,\mathrm{p,\,min}}$. %
The upscaled normal map $\bar{N}^l$ is used by one of the proposed extensions to account for non-fronto-parallel surfaces in the \gls*{SGM} optimization. %
In this, we reinitialize the subsequent steps and compute the depth, confidence and normal map corresponding to the next pyramid level. %
We denote the final depth, confidence and normal map, which are predicted at the lowest and finest pyramid level, as $\hat{D}$, $\hat{C}$ and $\hat{N}$, respectively.

A final \gls*{DoG} filter \citep{Wenzel2016dense} is used to unmask image regions, which do not provide enough texture to perform a reliable matching.

\subsection{Semi-Global Matching}
\label{sec:sgm}
\glsreset{SGM}

The \gls*{SGM} algorithm \citep{Hirschmueller2005, Hirschmueller2008} uses dynamic programming to efficiently approximate energy minimization of a two-dimensional \gls*{MRF} by independently aggregating the matching costs along numerous concentric one-dimensional paths. %
Along each path of direction $\mathrm{r}$, \gls*{SGM} recursively aggregates the matching costs $L_\mathrm{r}(\mathrm{p}, s)$ for a given pixel $\mathrm{p}$ and disparity $s \in \mathcal{T}=\{s_{min}, ... , s_{max}\}$ according to %
\begin{equation}
\label{eq:sgm-path}
\begin{aligned}
	L_\mathrm{r}(\mathrm{p}, s) = \mathcal{S}(\mathrm{p}, s) + \min\limits_{s'\in \mathcal{T}}\left(L_\mathrm{r}(\mathrm{p-r}, s')+\mathcal{V}(s,s')\right).
\end{aligned}
\end{equation}
Here, $\mathcal{S}(\mathrm{p}, s)$ denotes the unary data term, holding the matching cost stored inside the cost volume $\mathcal{S}$, while $\mathcal{V}(s, s')$ represents a smoothness term that penalizes deviations in the disparity $s$ of the pixel $\mathrm{p}$ and the disparity $s'$ of a neighboring pixel to $\mathrm{p}$ along the path, \ie the disparity of the previously considered pixel: % 
\begin{equation}
\label{eq:sgm-smooth}
\begin{aligned}
\mathcal{V}(s,s') &= 
	\begin{cases}
			0 &,\ \text{if}\ s = s' \\
    	P_1 &,\ \text{if}\ \left|s-s'\right| = 1 \\
        P_2 &,\ \text{if}\ \left|s-s'\right| > 1.
    \end{cases}
\end{aligned}
\end{equation}
At each pixel, the individual path costs are summed up, resulting in an aggregated cost volume 
\begin{equation}
\label{eq:sgm-aggr}
\begin{aligned}
\bar{\mathcal{S}}(\mathrm{p}, s) &= \sum\limits_r{L_\mathrm{r}(\mathrm{p}, s)}
\end{aligned}
\end{equation}
from which the pixel-wise winning disparities are extracted according to
\begin{equation}
\label{eq:sgm-wta}
\begin{aligned}
S(\mathrm{p}) = \argmin\limits_s{\bar{\mathcal{S}}(\mathrm{p}, s)}.
\end{aligned}
\end{equation}

\subsection{Extensions of the Semi-Global Matching Algorithm (\xSGM)}
\label{sec:xsgm}

\begin{figure*}[ht!]
	\centering
	\includegraphics[width=\textwidth]{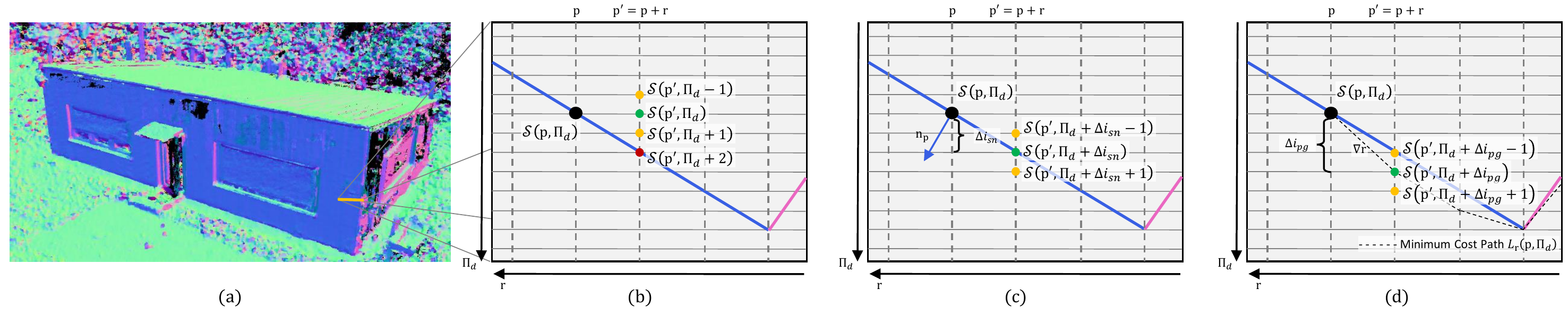}	
	\caption{\footnotesize{Illustration of the \xSGM path aggregation along one path direction $\mathrm{r}$. (a) Normal map of a building. The yellow line indicates the area for which the \xSGM path aggregation is shown. (b) Illustration of the \fpSGM path aggregation. The blue and pink lines correspond to blue and pink surface orientations on the building facade. When aggregating the path costs for pixel $\mathrm{p}$ at plane $\Pi_d$, \fpSGM will incorporate the previous costs at the same plane position (green) without additional penalty. The previous path costs at $\Pi_d\,\pm 1$ (yellow) will be penalized with $P_1$. The previous path costs located at $\Pi_d+2$ (red), which is actually located on the corresponding surface, will be penalized with the highest penalty $P_2$. (c) \snSGM uses the normal vector $\mathrm{n_p}$, encoding the surface orientation at pixel $\mathrm{p}$, and computes a discrete index jump $\Delta i_{sn}$, which adjusts the zero cost transition, causing the previous path costs at $\Pi_d+2$ to not be penalized. (d) Similar to \snSGM, \pgSGM adjusts the zero cost transition. However, the discrete index jump $\Delta i_{pg}$ is derived from the running gradient $\nabla \mathrm{r}$ of the minimum cost path. As illustrated, however, this can overcompensate the shift of the zero cost transition.}}	
	\label{fig:xsgm} 
\end{figure*}
	
The first of the proposed \xSGM extensions, which at the same time serves as a basis to the other two extensions, is a straight-forward adaptation of the standard \gls*{SGM} approach to the use of a fronto-parallel multi-view plane-sweep sampling as part of the work flow presented in \Cref{fig:methodology}. %
It is thus denoted as \textit{fronto-parallel} \gls*{SGM} (\fpSGM) and was already used in \citep{Ruf2017cross} and \citep{Ruf2018deep}. %
The recursive aggregation of the matching costs along each path is adjusted to %
\begin{equation}
\label{eq:xsgm}
\begin{aligned}
	L_\mathrm{r}(\mathrm{p}, \Pi_d) =\  &\mathcal{S}(\mathrm{p}, \Pi_d)\ +\ \\ 
	\ &\min\limits_{d'\in \mathcal{D}}\left(L_\mathrm{r}(\mathrm{p-r}, \Pi_{d'})+\mathcal{V}_{fp}(\Pi_d,\Pi_{d'})\right),
\end{aligned}
\end{equation}
with $\Pi_d$ being the sampling plane at depth $d$ used to perform the multi-image matching. %
Here, instead of penalizing the deviations in neighboring disparities, the smoothness term $\mathcal{V}_{fp}$ penalizes different planes between adjacent pixels along the path $L_\mathrm{r}$:%
\begin{equation}
\label{eq:sgm-smooth-fp}
\begin{aligned}
\mathcal{V}_{fp}(\Pi_d,\Pi_{d'}) &= 
	\begin{cases}
			0 &,\ \text{if}\ \Gamma(\Pi_d) = \Gamma(\Pi_{d'}) \\
    	P_1 &,\ \text{if}\ \left|\Gamma(\Pi_d) - \Gamma(\Pi_{d'})\right| = 1 \\
        P_2 &,\ \text{if}\ \left|\Gamma(\Pi_d) - \Gamma(\Pi_{d'})\right| > 1,
    \end{cases}
\end{aligned}
\end{equation}
with $\Gamma(\cdot)$ being a function that returns the index of $\Pi_d$ within the set of sampling planes (\cf \Cref{fig:xsgm}(b)).%

%As already stated, \fpSGM still favors fronto-parallel surfaces. %
In our second extension, namely \textit{surface normal} \gls*{SGM} (\snSGM), we adopt the approach presented by \citet{Scharstein2018surface} to use surface normals to adjust the zero-cost transition to coincide with the surface orientation. %
We use a normal map $N$ corresponding to $I_\mathrm{ref}$ that holds the surface normals of the scene that is to be reconstructed. %
We assume $N$ to be given or use the normal map that has been predicted in the previous iteration of our hierarchical work flow (\cf \Cref{fig:methodology}). %
Assuming that a surface normal vector $\mathrm{n}_\mathrm{p} = N(\mathrm{p})$ corresponding to the surface orientation at pixel $\mathrm{p}$ is given,
we compute the discrete index jump ${\Delta i}_{sn}$ through the set of sampling planes that is caused by the tangent plane to $\mathrm{n}_\mathrm{p}$ at the scene point $\mathrm{X_p}$, in which the ray through $\mathrm{p}$ intersects $\Pi_d$. %
As also stated by \citet{Scharstein2018surface}, these discrete index jumps can be computed once for each pixel $\mathrm{p}$ and each path direction $\mathrm{r}$ (\cf \Cref{fig:xsgm}(c)). %
Given ${\Delta i}_{sn}$, we adjust the smoothness term used in \snSGM according to 
\begin{equation}
\label{eq:sgm-smooth-sn}
\begin{aligned}
\mathcal{V}_{sn}(\Pi_d,\Pi_{d'}) = \mathcal{V}_{fp}(\Pi_d + {\Delta i}_{sn},\Pi_{d'}).
\end{aligned}
\end{equation}

The third extension does not consider any additional information, such as surface normals $N$, while computing the aggregating path costs $L_\mathrm{r}(\mathrm{p}, \Pi_d)$. %
Instead, it relies on the gradient $\nabla \mathrm{r}$ in scene space corresponding to the minimal path costs and is therefore denoted as \textit{path gradient} \gls*{SGM} (\pgSGM). %
Here, we again consider $\mathrm{X_p}$ as the scene point corresponding to the intersection between the ray through $\mathrm{p}$ and $\Pi_d$. %
Furthermore, we denote $\mathrm{p'} = \mathrm{p+r}$ as predecessor of $\mathrm{p}$ along the path $\mathrm{r}$ and $\mathrm{\hat{X}_{p'}}$ as the scene point parameterized by $\mathrm{p'}$ and the plane $\Pi_{\hat{d}}$. %
The latter represents the plane at depth $\hat{d} = \argmin_{d'\in \mathcal{D}}{L_\mathrm{r}(\mathrm{p'}, \Pi_{d'})}$ associated with the previous minimal path costs. %
From this, we dynamically compute a gradient vector $\nabla \mathrm{r} = \mathrm{X_p} - \mathrm{\hat{X}_{p'}}$ in scene space while traversing along the path $\mathrm{r}$. %
Given $\nabla \mathrm{r}$, we again compute a discrete index jump ${\Delta i}_{pg}$ through the set of sampling planes and use 
\begin{equation}
\label{eq:sgm-smooth-pg}
\begin{aligned}
\mathcal{V}_{pg}(\Pi_d,\Pi_{d'}) = \mathcal{V}_{fp}(\Pi_d + {\Delta i}_{pg},\Pi_{d'})
\end{aligned}
\end{equation}
to dynamically adjust the zero-cost transition to possibly slanted surfaces in scene space (\cf \Cref{fig:xsgm}(d)). %
This allows us to implicitly penalize deviations from the running gradient vector in scene space between two consecutive pixels along the path $\mathrm{r}$. %

Since our extensions \xSGM only affect the path-wise aggregation of the matching costs, we extract the depth map $D$ analogously to \Cref{eq:sgm-aggr} and \Cref{eq:sgm-wta}, substituting the disparity by the depths corresponding to the set of sampling planes. %
Note that, since our sampling set consists of fronto-parallel planes, we can directly extract the depth from their parameterization. %
If slanted planes are used for sampling, a pixel-wise intersection of the viewing rays with the winning planes is to be performed in order to extract $D$. %

Finally, for each of the extensions, a median filter with a kernel size of $5\times 5$ is used to further reduce noise. % 

\subsection{Adaptive smoothness penalty}

\citet{Hirschmueller2005, Hirschmueller2008} suggests to adaptively adjust the penalty $P_2$ to the image gradient along path $\mathrm{r}$ in order to preserve depth discontinuities at object boundaries. %
In this work, we evaluate two different strategies to adjust $P_2$. % 
The first strategy fully relies on the absolute intensity difference ($|\Delta I|$) between consecutive pixels:
\begin{equation}
\label{eq:p2-grad}
\begin{aligned}
P_2^{\Delta I} = P_1\left(1+\alpha \exp\left(-\frac{|\Delta I|}{\beta}\right)\right)
\end{aligned}
\end{equation}
with $\alpha=8$ and $\beta=10$ according to \citet{Scharstein2018surface}.

A second strategy that has been proposed by \citet{Ruf2018deep} relies on the use of a line segment detector \citep{Gioi2010lsd} to generate a binary line image of the reference image $I_{\mathrm{ref}}^{line}$ and reduce $P_2^{line}$ to $P_1$ at a detected line segment:
\begin{equation}
\label{eq:p2-line}
\begin{aligned}
P_2^{line} &= 
	\begin{cases}
    	P_1 &,\ \text{if}\ I^{line}_\mathrm{ref}(\mathrm{p}) = 1 \\
        P_2 &,\ \text{otherwise}.
    \end{cases}
\end{aligned}
\end{equation}
\citet{Ruf2018deep} argue that this allows to enforce strong discontinuities at object boundaries while increasing the smoothness within objects.

\subsection{Confidence measure}
\label{sec:conf}

We additionally compute a confidence map $C$, holding per-pixel confidence measures of the depth estimates in the range of $\left[0,1\right]$. %
For this, we model two confidence measures that solely rely on the results of the \gls*{SGM} path aggregation. %
With our first measure $U_{p}$, we adopt the observation of \citet{Drory2014semi}, that the sum of the  individual minimal path costs at pixel $\mathrm{p}$ is a lower bound of the winning aggregated costs:
\begin{equation}
\label{eq:conf-path}
\begin{aligned}
U_{p} &= \min\limits_{d}{\bar{\mathcal{S}}(\mathrm{p}, \Pi_d)} - \sum\limits_\mathrm{r}{\min\limits_{d}{L_\mathrm{r}(\mathrm{p}, \Pi_d)}}.
\end{aligned}
\end{equation}
The second confidence measure $U_{u}$ models the uniqueness of the winning aggregated costs, \ie the difference between the lowest and second-lowest aggregated costs for each pixel in $\bar{\mathcal{S}}$:
\begin{equation}
\label{eq:conf-unique}
\begin{aligned}
U_{u} = &\min\limits_{d}\left(\bar{\mathcal{S}}(\mathrm{p}, \Pi_d) \backslash \min\limits_{d}{\bar{\mathcal{S}}(\mathrm{p}, \Pi_d)}\right) -\\ &\min\limits_{d}{\bar{\mathcal{S}}(\mathrm{p}, \Pi_d)}.
\end{aligned}
\end{equation}
Given the above measures, we compute the final pixel-wise confidence value according to
\begin{equation}
\label{eq:conf}
\begin{aligned}
C(\mathrm{p}) = \exp\left(- \frac{U_{p}}{\varphi}\right) \cdot \min\left\{\exp\left(U_{u} - \tau\right), 1\right\}.
\end{aligned}
\end{equation}
In this equation, the first term will resolve to 1, if $U_p=0$, \ie the winning costs equal the sum of the minimal path costs. %
If this is not the case, the rate of the exponential decay of the confidence is controlled by the parameter $\varphi$. %

The parameter $\tau$ represents the uniqueness threshold of the winning solution. %
If the absolute difference between the lowest and second-lowest pixel-wise aggregated costs in $\bar{\mathcal{S}}$ is above the threshold $\tau$, the second term of \Cref{eq:conf} will resolve to 1.%

In our evaluation, this confidence measure is used to plot the accuracy of the predicted depth maps with respect to their completeness, where the latter is computed by thresholding the corresponding confidence map (\cf \Cref{fig:quant_results}(b)). %

\subsection{Normal map estimation}
\label{sec:normalmap}

The third output of our algorithm is a normal map $N$ that holds the surface orientation in the depth map $D$ at pixel $\mathrm{p}$. %
For the computation of the surface normal, we reproject the depth map into a point cloud and compute the cross product $\mathrm{n_p} = \mathrm{h_p}\times \mathrm{v_p}$. %
Here, $\mathrm{h_p}$ denotes a vector between the scene points of two neighboring pixels to $\mathrm{p}$ in horizontal direction, while $\mathrm{v_p}$ is the vector between the scene points of two vertical neighboring pixels. %

Since the computation of $N$ does not contain any smoothness assumption, we apply an a-posteriori smoothing to the normal map, the so-called Gestalt-Smoothing. %
In particular, we perform an appearance-based weighted Gaussian smoothing in a local two-dimensional neighborhood $\mathcal{N}_\mathrm{p}$ around $\mathrm{p}$:
\begin{equation}
\label{eq:normal}
\begin{aligned}
N(\mathrm{p}) = \frac{\bar{\mathrm{n}}_\mathrm{p}}{\left|\bar{\mathrm{n}}_\mathrm{p}\right|},
\end{aligned}
\end{equation}
with
\begin{equation}
\label{eq:normal-smooth}
\begin{aligned}
\bar{\mathrm{n}}_\mathrm{p} = \mathrm{n_p} + \sum\limits_{\mathrm{q}\in\mathcal{N}_\mathrm{p}}\left[\mathrm{n_q} \cdot \frac{1}{\sqrt{2\pi\sigma^2}}\exp\left(-\frac{\left(\mathrm{q}-\mathrm{p}\right)^2}{2\sigma^2}\right)\right. \\ \left. \cdot \exp\left(-\frac{|I_\mathrm{q}-I_\mathrm{p}|}{\beta}\right)\right],
\end{aligned}
\end{equation}
where $\beta = 10$ in accordance with \Cref{eq:p2-grad}, and $\sigma$ is fixed to the radius of the local smoothing neighborhood.

%%%%%%%%%%%%%%%%%%%%%%%%%%%%%%%%%%%%%%%%%%%%%%%
\section{EVALUATION}
\label{sec:eval}
%%%%%%%%%%%%%%%%%%%%%%%%%%%%%%%%%%%%%%%%%%%%%%%
% KAO: Sloppy spacing ensures non-overfull lines. Can be removed if this is not an issue.
\sloppy
\glsreset{NCC}
\glsreset{CT}

%\subsection{Datasets}

\subsection{Experiments}
\label{sec:experiments}

% results without agnular error
\renewcommand{\arraystretch}{1.5}
\begin{figure*}[ht!]
	\subfigure[Quantitative results of all twelve configurations which are evaluated on the DTU and TMB dataset. For each dataset, the mean absolute L1 error (mL1-abs) as well as the mean relative L1 error (mL1-rel) are evaluated. The configuration name encodes the different configuration settings. Here, the first part represents the extension used, the middle section holds the cost function which was applied in the multi-image matching, and the third portion represents the strategy, which was adopted to adapt the $P_2$ penalty. The last row denotes the results achieved by the offline \gls*{SFM} pipeline COLMAP \citep{Schoenberger2016mvs, Schoenberger2016sfm}.]{
		\centering
		\scriptsize
		\begin{tabular}{|l|r|r|r|r|}
		\cline{2-5}
 		\multicolumn{1}{c|}{} &\multicolumn{ 2}{c|}{DTU} & \multicolumn{ 2}{c|}{TMB} \\
        %\textcolor{cyan}{
		\hline
	 	Configuration Name	& mL1-abs & mL1-rel  & mL1-abs & mL1-rel  \\ 
				\hline
	 	\fpSGM -CT-$P_2^{\Delta I}$ & 10.362\,\tiny{$\pm$11.867} & 0.014\,\tiny{$\pm$0.015} & 0.392\,\tiny{$\pm$0.336} & 0.712\,\tiny{$\pm$0.486} \\
				\hline
		 \fpSGM -CT-$P_2^{line}$ & 10.392\,\tiny{$\pm$12.065} & 0.014\,\tiny{$\pm$0.015} & 0.406\,\tiny{$\pm$0.358} & 0.713\,\tiny{$\pm$0.485} \\
				\hline
		 \fpSGM -NCC-$P_2^{\Delta I}$ & \underline{9.859}\,\tiny{$\pm$11.781} & \underline{0.014}\,\tiny{$\pm$0.015} & \underline{0.406}\,\tiny{$\pm$0.347} & \underline{0.704}\,\tiny{$\pm$0.480} \\
				\hline
		 \fpSGM -NCC-$P_2^{line}$ & 12.588\,\tiny{$\pm$13.493} & 0.017\,\tiny{$\pm$0.017} & 0.492\,\tiny{$\pm$0.435} & 0.704\,\tiny{$\pm$0.461} \\
				\hline
				\hline
		 \snSGM -CT-$P_2^{\Delta I}$ & 10.106\,\tiny{$\pm$11.532} & 0.014\,\tiny{$\pm$0.015} & 0.401\,\tiny{$\pm$0.349} & 0.717\,\tiny{$\pm$0.489} \\
				\hline
		 \snSGM -CT-$P_2^{line}$ & 10.292\,\tiny{$\pm$12.068} & 0.014\,\tiny{$\pm$0.016} & 0.412\,\tiny{$\pm$0.367} & 0.718\,\tiny{$\pm$0.489} \\
				\hline
		 \snSGM -NCC-$P_2^{\Delta I}$ & \underline{9.770}\,\tiny{$\pm$11.850} & \underline{0.013}\,\tiny{$\pm$0.015} & \underline{0.411}\,\tiny{$\pm$0.351} & \underline{0.705}\,\tiny{$\pm$0.479}  \\
				\hline
		 \snSGM -NCC-$P_2^{line}$ & 12.402\,\tiny{$\pm$13.405} & 0.017\,\tiny{$\pm$0.017} & 0.491\,\tiny{$\pm$0.434} & 0.704\,\tiny{$\pm$0.460} \\
				\hline
				\hline
		 \pgSGM -CT-$P_2^{\Delta I}$ & 10.612\,\tiny{$\pm$11.919} & 0.015\,\tiny{$\pm$0.015} & 0.401\,\tiny{$\pm$0.339} & 0.718\,\tiny{$\pm$0.493} \\
				\hline
		 \pgSGM -CT-$P_2^{line}$ & 10.529\,\tiny{$\pm$12.014} & 0.015\,\tiny{$\pm$0.015} & 0.413\,\tiny{$\pm$0.359} & 0.712\,\tiny{$\pm$0.492} \\
				\hline
		 \pgSGM -NCC-$P_2^{\Delta I}$ & \underline{10.010}\,\tiny{$\pm$11.739} & \underline{0.014}\,\tiny{$\pm$0.015} & \underline{0.417}\,\tiny{$\pm$0.344} & \underline{0.710}\,\tiny{$\pm$0.481} \\
				\hline
		 \pgSGM -NCC-$P_2^{line}$ & 12.598\,\tiny{$\pm$13.358} & 0.017\,\tiny{$\pm$0.017} & 0.495\,\tiny{$\pm$0.432} & 0.706\,\tiny{$\pm$0.461} \\
				\hline 
				\hline
		 COLMAP & 3.309\,\tiny{$\pm$4.156} & 0.005\,\tiny{$\pm$0.006} & - & - \\
		 		\hline
		\multicolumn{5}{c}{}
		%} %textcolor
		\end{tabular} 
	} \hfill 
	\subfigure[ROC curves plotting normalized mean L1-rel over the confidence threshold which is used to mask the depth map. The top graph depicts the results achieved by three \xSGM configurations on the DTU dataset. The bottom graph shows the results achieved on the TMB dataset.] { 
		\centering
		\begin{large}
		\raisebox{-.480\height}{\resizebox{0.8\columnwidth}{!}{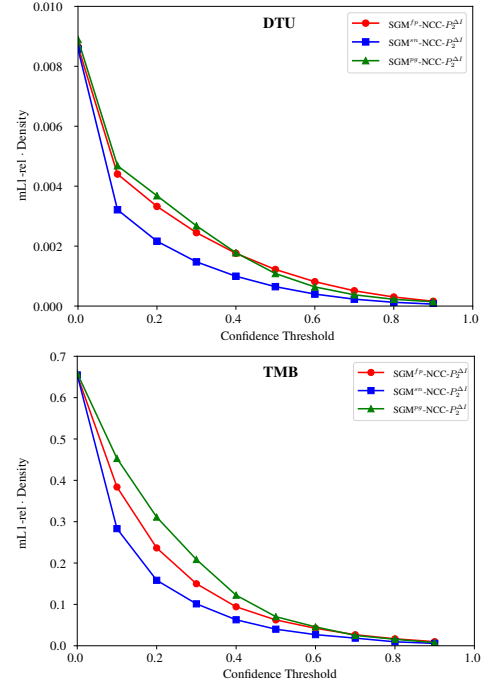}}
		\end{large}
	}
\caption{Quantitative evaluation of twelve different \xSGM configurations.}
\label{fig:quant_results}  
\end{figure*}

We have evaluated our approach on two different datasets, namely the DTU Robot \gls*{MVS} dataset \citep{Jensen2014dtu} and a private dataset, which is henceforth referred to as the TMB dataset. %

From the DTU dataset, we have selected 21 scans of the different building models, in which each model was captured from 49 locations and with eight different lighting conditions. %
For our evaluation, we have used the already undistorted images with a resolution of 1600$\times$1200 pixels, captured under the most diffuse lighting. %
As ground truth to our approach, we have extracted depth maps from the structured light scans, which are included in the dataset, given the camera poses of the reference image.

Our privately captured TMB dataset consists of three different scenes captured with a DJI Phantom 3 Professional from multiple different aerial viewpoints. %
The images were captured while flying around the objects of interest at three different altitudes (8\m, 10\m and 15\m). %
Each image was resized to 1920$\times$1080 pixels before used for our evaluation.
We compare the results achieved by the proposed approach to data from an offline \gls*{SFM} pipeline for accurate and dense 3D image matching. %
For this, we have used COLMAP \citep{Schoenberger2016mvs, Schoenberger2016sfm} to reconstruct the considered scenes of the dataset. %
As a required input to our algorithm, we have used the camera poses computed by the sparse reconstruction. %
In the evaluation, we have compared the depth maps predicted by our algorithm against the geometric depth maps from the dense reconstruction of COLMAP. %
%Here, we have used the pinhole camera model in case of the already undistorted imagery from the DTU dataset. %
%For the images from the TMB dataset we have selected the OpenCV parameterization based on the distortion model proposed by \citet{Brown1971calibration}. %

%The photometric normal maps are chosen over the geometric normal maps due to the fact, that they hold the surface orientation with respect to the camera center. %
%However, since the photometric normal maps are not yet filtered from noise we only consider the surface normals at pixels at which a depth is provided in the corresponding geometric depth map. %
%
%We are not able to directly compare the results achieved by our approach with the above mentioned work, since the above mentioned work focus on the use of the \gls*{SGM} for a two-view stereo reconstruction and are evaluated only on corresponding benchmarks. %

As an accuracy measure between the estimates and the ground truth, we have used an absolute and relative L1 measure, which is computed pixel-wise and averaged over the number of estimates in the depth map: %
\begin{equation}
\label{eq:l1abs-measure}
\begin{aligned}
\text{L1-abs}(d, \hat{d}) = \frac{1}{m}\sum\limits_i{|d_i - \hat{d}_i|},
\end{aligned}
\end{equation}
\begin{equation}
\label{eq:l1rel-measure}
\begin{aligned}
\text{L1-rel}(d, \hat{d}) = \frac{1}{m}\sum\limits_i{\frac{|d_i - \hat{d}_i|}{\hat{d}_i}},
\end{aligned}
\end{equation}
with $d$ and $\hat{d}$ denoting the predicted and ground truth depth values respectively, and with $m$ being the number of pixels for which both $d$ and $\hat{d}$ exists. %
Here, the depth values denote the fronto-parallel distances of the corresponding scene points from the image center of the reference camera.
Both measures are only evaluated for pixels which contain a predicted and ground truth depth value. %
While L1-abs denotes the average absolute difference between the prediction and the ground truth, L1-rel computes the depth error relative to the ground truth depth.
This reduces the influence of a high absolute error where the ground truth depth is large and increases the influence of measurements close to the camera.
This is important, as the uncertainty of the depth measurements typically increases with the distance from the camera. %

%For the quantitative evaluation of the normal maps, we have used an pixel-wise averaged absolute angular error between the computed normal vector $\mathrm{n}$ and the ground truth vector $\hat{\mathrm{n}}$:
%%
%\begin{equation}
%\label{eq:aae-measure}
%\begin{aligned}
%\text{AAE}(\mathrm{n}, \hat{\mathrm{n}}) = \frac{1}{m}\sum\limits_i{\left|\Phi\left(\mathrm{n}_i, \hat{\mathrm{n}}_i\right)\right|},
%\end{aligned}
%\end{equation}
%%
%with $\Phi(\cdot)$ computing the enclosing angle between the two vectors, and with $m$ again being the number of pixels for which both $\mathrm{n}$ and $\hat{\mathrm{n}}$ exists. %

The parameterization of our algorithms was determined empirically based on the results, which were achieved on the DTU dataset. %
For our hierarchical processing scheme, we have used $L=3$ pyramid levels and have set $\Delta d = 6$ for the computation of the per-pixel sampling range, yielding the best trade-off between accuracy and runtime. %
The support region of the \gls*{NCC} and the \gls*{CT} was set to 5$\times$5 and 9$\times$7, respectively, where the latter is the maximum size for which the \gls*{CT} bit string still fits into a 64 bit integer. %
In the computation of the normal map, we have used a smoothing kernel of size 21$\times$21 for the Gestalt-Smoothing. %

Due to the different range of values in the cost functions, the parameterization of the penalty functions and the confidence measures need to be chosen accordingly. %
For the \gls*{NCC} similarity measure, we have set $P_1 = 150$ when using $P_2^{\Delta I}$ and $P_1 = 60,P_2=220$ when using $P_2^{line}$. %
In case of the \gls*{CT}, we have set $P_1 = 15$ when using $P_2^{\Delta I}$ and $P_1 = 10, P_2=55$ when using $P_2^{line}$. %
For the computation of the confidence measure (cf. \Cref{eq:conf}), we have set $\varphi=80, \tau=10$ when using the \gls*{NCC}, and $\varphi=650, \tau=80$ when using the \gls*{CT}. %

\begin{figure*}[ht!]
	\centering
	\includegraphics[width=\textwidth]{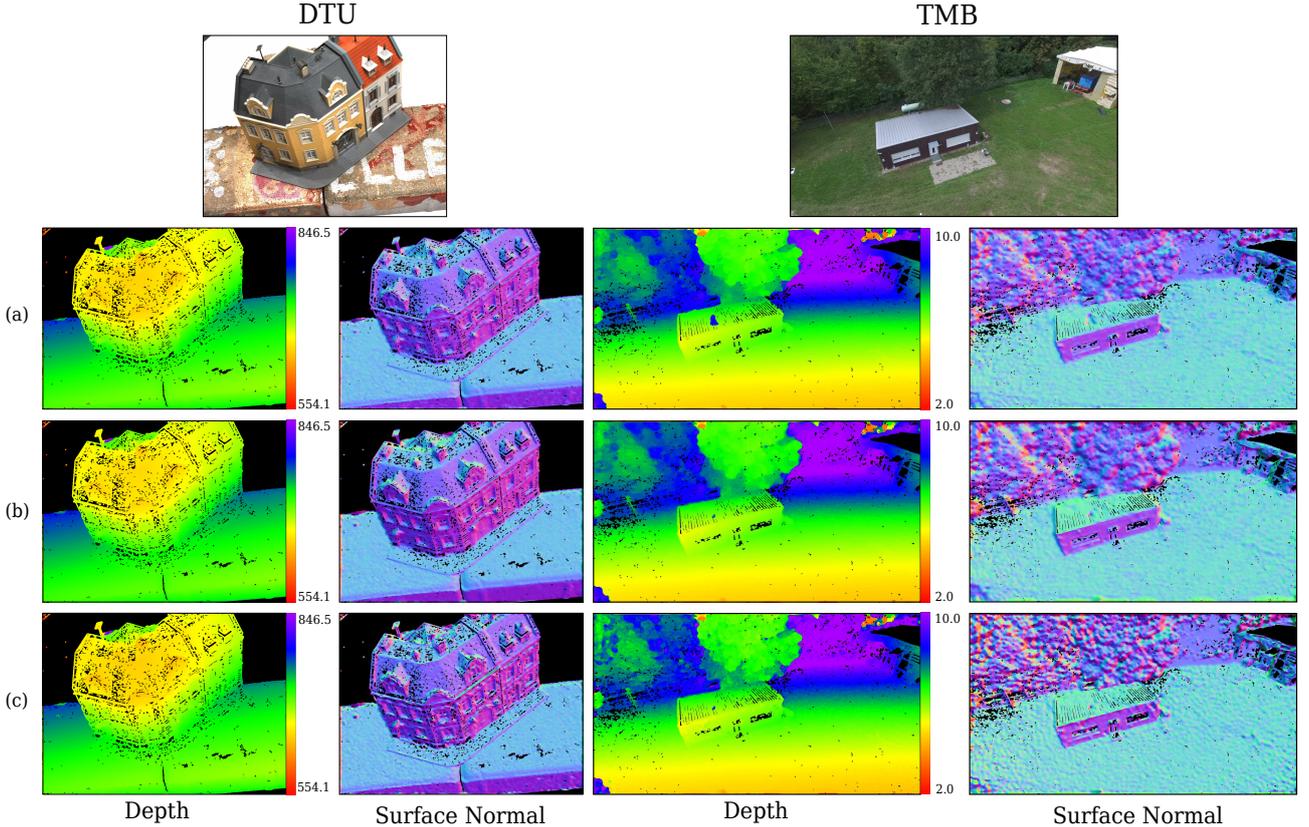}
	\caption{\footnotesize{Qualitative comparison between the results achieved by the \fpSGM -NCC-$P_2^{\Delta I}$ (Row (a)), \snSGM -NCC-$P_2^{\Delta I}$ (Row (b)) and \pgSGM -NCC-$P_2^{\Delta I}$ (Row (c)). Note that the depth and normal maps are filtered by the \gls*{DoG} filter and according to the available data in the ground truth.}}
	\label{fig:qual_results} 
\end{figure*}

All experiments were performed on a desktop hardware with an Intel Core i7-5820K CPU \@ 3.3GHz and a NVIDIA GeForce GTX 1070 GPU. %
The computationally expensive part of our algorithm, such as depth and normal map estimation, is optimized with CUDA to run on the GPU, achieving a frame rate of 1Hz$-$2Hz for full HD imagery, depending on the configuration and parameterization used. %

In the scope of this work, we have evaluated twelve different configurations of our \xSGM extensions. %
The results achieved by these configurations on the two datasets are listed in \Cref{fig:quant_results}(a). %
The configuration names denote the corresponding setups. %
In comparison, the results achieved by the offline \gls*{SFM} pipeline COLMAP are listed in the last row of \Cref{fig:quant_results}(a). %
The values in the depth maps are in the range of $\left[554.1, 846.5\right]$ in case of the DTU dataset, and $\left[2.0, 10\right]$ in the depth maps corresponding to the TMB dataset (\cf \Cref{fig:qual_results}). %
Since the datasets do not have any metric system, the errors are without any unit. %
However, the given ranges of values in the depth maps allow to draw conclusions on these error values with respect to the estimates. %

For each of the three \xSGM extensions, we have chosen one configuration, namely the one with the lowest average error (underlined values), and plotted ROC curves for both datasets (\cf \Cref{fig:quant_results}(b)). %
These curves represent the mean relative L1 error, normalized by the density, over the confidence threshold, which is used to mask the corresponding depth map. %
A qualitative comparison between the three different \xSGM extensions is done on one example image from each of the two datasets (\cf \Cref{fig:qual_results}). %
Here, the depth and normal maps are filtered by the \gls*{DoG} filter and according to available data in the ground truth. %
A discussion of the experimental results is given in the section below. %

\subsection{Discussion of the experimental results}
\label{sec:discussion}

% KAO: Sloppy spacing ensures non-overfull lines. Can be removed if this is not an issue.
\sloppy 

The experimental results listed in \Cref{fig:quant_results} reveal a number of strengths and weaknesses of the proposed \xSGM extensions. %
Starting off with the strengths, the numbers in \Cref{fig:quant_results}(a) show similar accuracies between all three extensions. %
Especially when considering the value ranges in the depth maps one can argue that, with an absolute error of 1\%$-$4\% of the maximum depth range, all \xSGM approaches achieve a high accuracy with respect to the ground truth. %
The differences in accuracy between the two datasets can be attributed to the fact that the parameters of our approach were fine-tuned with respect to the DTU dataset. %
Furthermore, since the ranges of depth values in the TMB dataset are much smaller than the ones in the DTU dataset, a minor difference between the estimate and the ground truth has a greater influence on the resulting error, which opposes the implication of a possible parametric over-fitting with respect to the DTU dataset. %

A comparison between the three \xSGM extensions, solely based on the values listed in \Cref{fig:quant_results}(a), does not allow to conclude that the consideration of surface normals significantly increases the accuracy of the resulting depth map. %
In fact, looking at the results from the TMB dataset, the mean errors achieved by \fpSGM are lower than the ones achieved by \snSGM and \pgSGM. %
However, the ROC curves in \Cref{fig:quant_results}(b) reveal that the use of surface normals increases the ratio between the accuracy of the measurements and their confidence. %
%While in case of the TMB dataset only the curve of \snSGM is below the one of \fpSGM, the curves corresponding to the DTU dataset suggest a superiority of both \snSGM and \pgSGM over \fpSGM. %
This assumption is also encouraged by a qualitative comparison based on the results depicted in \Cref{fig:qual_results}.
The use of surface normals yields a slightly more consistent normal map, in particular when comparing the roof of the buildings. %
This supports the claims of \cite{Scharstein2018surface}. %
%While the surface orientation of the building roof in the normal map of \fpSGM (\cf \Cref{fig:qual_results}(a)) is not completely uniform, the results of \snSGM (\cf \Cref{fig:qual_results}(b)) and \pgSGM (\cf \Cref{fig:qual_results}(c)) show that the normal maps are more consistent in homogeneous areas. %

However, the qualitative comparison also reveals that the depth discontinuities at object boundaries are less concise when considering surface normals in the \gls*{SGM} optimization. %
While this could result from the adjustment of the zero transition in the path aggregation, it cannot be ruled out that this is due to less appropriate parameterization of the penalty functions. %
Furthermore, the evaluation reveals that the results achieved by \pgSGM are inferior to the ones of \fpSGM and \snSGM as the normal maps are more noisy compared to the other results. %
While a cause of this effect could not be fully resolved in the scope of this work, we believe that this might be attributed to an overcompensation in the extraction of the zero transition shift from the gradient of the minimal cost path.%

An evaluation of the different cost functions and different penalty configurations used has not revealed a clear winner. %
In fact, it depends on the nature of the dataset and the parameterization of the algorithm. % 
Nonetheless, the values in \Cref{fig:quant_results}(a) suggest that, in most cases, the use of \gls*{NCC} in the image matching and $P_2^{\Delta I}$ in the \gls*{SGM} optimization is an appropriate choice. % 

Lastly, in this work, we have only considered the use of fronto-parallel plane orientation while performing the plane-sweep multi-image matching. %
Yet, \citet{Gallup2007} and \citet{Sinha2014} suggest to use multiple sweeping directions for the plane-sweep sampling. %
Doing so would not require to incorporate surface orientations in the \gls*{SGM} optimization, but allow to use the standard \gls*{SGM} (\cf \fpSGM) in its adaption to plane-sweep stereo as done by \cite{Sinha2014}. %
An evaluation of this is an interesting direction for future work.

\section{CONCLUSION \& FUTURE WORK}
\label{sec:conclusion}

% KAO: Sloppy spacing ensures non-overfull lines. Can be removed if this is not an issue.
\sloppy

In conclusion, this work proposes a hierarchical algorithm for efficient depth and normal map estimation from oblique aerial imagery based on plane-sweep multi-image matching followed by a semi-global matching optimization for cost aggregation and regularization. %
Our approach allows to additionally consider local surface orientations in the computation of the depth map. %

Both the standard \gls*{SGM} optimization and the adjustment of the same with respect to local surface normals, achieve results with high accuracies. %
However, our experiments support the claims of \cite{Scharstein2018surface} that the consideration of surface normals achieves more consistent results with higher confidence in homogeneous areas. %
Furthermore, the quantitative evaluation reveals that our results are comparable to the ones achieved by sophisticated \gls*{SFM} pipelines such as COLMAP. %
In contrast, however, our approach only considers a confined image bundle of an input sequence allowing to perform an online computation at 1Hz$-$2Hz. %
%Thus, our approach allows for an online structural scene analysis from oblique aerial imagery, if joined with the online estimation of camera poses and simultaneous localization with respect to a reference system.

Nonetheless, the experimental results have also revealed a number of improvements and considerations that are promising options for future work. %
An example is the mentioned incorporation of multiple plane orientations in the process of plane-sweep multi-image matching. %
Another aspect, which is to be considered in future work, is the computation and evaluation of the normal map. %
We have extracted the normal map solely based on the geometric information in the depth map with an a-posteriori smoothing. %
However, the use of a more sophisticated method would greatly improve the results. %

{\footnotesize % tune the size by altering the parameter
	\begin{spacing}{0.7}% tune the spacing between entries
    \setlength{\bibsep}{1pt}
		\bibliography{multiview-gradient-sgm_biblio} % Include your own bibliography (*.bib), style is given in isprs.cls
	\end{spacing}
}

\end{document}